\pdfoutput=1

\documentclass[11pt]{article}

\usepackage[]{acl}
\usepackage{times}
\usepackage{latexsym}
\usepackage{multicol}
\usepackage{multirow}
\usepackage{booktabs}
\usepackage{amsmath} 
\usepackage{amssymb}  
\usepackage{mdframed}
\usepackage{xcolor}
\usepackage{pifont} 
\usepackage{latexsym}
\usepackage{authblk}
\usepackage{tabularx}
\usepackage[T1]{fontenc}

\usepackage[utf8]{inputenc}

\usepackage{microtype}

\usepackage{inconsolata}

\usepackage{graphicx}

\title{Theory of Mind in Large Language Models: Assessment and Enhancement}

\author{\bf Ruirui Chen$^1$, Weifeng Jiang$^3$, Chengwei Qin$^{3, 4}$, Cheston Tan$^{1, 2}$\\
\textsuperscript{1}Institute of High Performance Computing (IHPC) and
\textsuperscript{2}Centre for Frontier AI Research (CFAR),\\
Agency for Science, Technology and Research (A*STAR), Singapore \\
\textsuperscript{3}Nanyang Technological University, Singapore \\
\textsuperscript{4}Hong Kong University of Science and Technology (Guangzhou), China 
}

\begin{document}
\maketitle
\begin{abstract}
Theory of Mind (ToM)—the ability to reason about the mental states of oneself and others—is a cornerstone of human social intelligence. As Large Language Models (LLMs) become increasingly integrated into daily life, understanding their ability to interpret and respond to human mental states is crucial for enabling effective interactions. In this paper, we review LLMs' ToM capabilities by analyzing both evaluation benchmarks and enhancement strategies. For evaluation, we focus on recently proposed and widely used story-based benchmarks. For enhancement, we provide an in-depth analysis of recent methods aimed at improving LLMs' ToM abilities. Furthermore, we outline promising directions for future research to further advance these capabilities and better adapt LLMs to more realistic and diverse scenarios. Our survey serves as a valuable resource for researchers interested in evaluating and advancing LLMs' ToM capabilities.
\end{abstract}

\section{Introduction}

Theory of Mind (ToM) refers to the ability to attribute mental states, such as emotions, intentions, and beliefs, to oneself and others \cite{Premack_Woodruff_1978}. In essence, ToM involves "putting yourself in someone else's shoes" \cite{goldman2006simulating, wilf-etal-2024-think}. It enables us to "mind read" and understand how others feel, what their intentions are, what they believe or know. This ability to infer what others believe in a given situation is crucial for predicting their actions, forming a fundamental component of social skills. Research has shown that the ability to attribute mental states to others begins developing around the second year of life, with a fully explicit ToM capability within reach by the age of four \cite{WIMMER1983103, BARONCOHEN198537}.

Recently, Large Language Models (LLMs) have achieved remarkable success across various tasks, particularly in Natural Language Processing (NLP) \cite{qin-etal-2023-chatgpt}.
Currently, there is growing interest to explore the cognitive abilities of LLMs through novel and intriguing tasks, including assessments of their Emotional Intelligence (EI) \cite{yang-etal-2023-towards, huang2024humanity, 10572294, sabour-etal-2024-emobench, liu2024emollms}, strategic reasoning capabilities \cite{gandhi2023strategic, zhang2024llm}, and ToM capabilities \cite{wu-etal-2023-hi, van-duijn-etal-2023-theory, wu-etal-2024-coke, shapira-etal-2024-clever, chen-etal-2024-tombench, jin-etal-2024-mmtom}. 
Among these capabilities, there is an ongoing debate about whether LLMs truly possess ToM abilities. Some studies suggest that LLMs demonstrate promising signs of ToM competence \cite{bubeck2023sparks, doi:10.1073/pnas.2405460121, street2024llms}, while others contend that these abilities are often superficial and unstable \cite{shapira-etal-2024-clever, ullman2023large, ma-etal-2023-tomchallenges, 10.1145/3610978.3640767, 10.5555/3692070.3694663, bortoletto-etal-2024-limits}. 
\begin{figure*}[t]
    \centering
    \includegraphics[width=2\columnwidth]{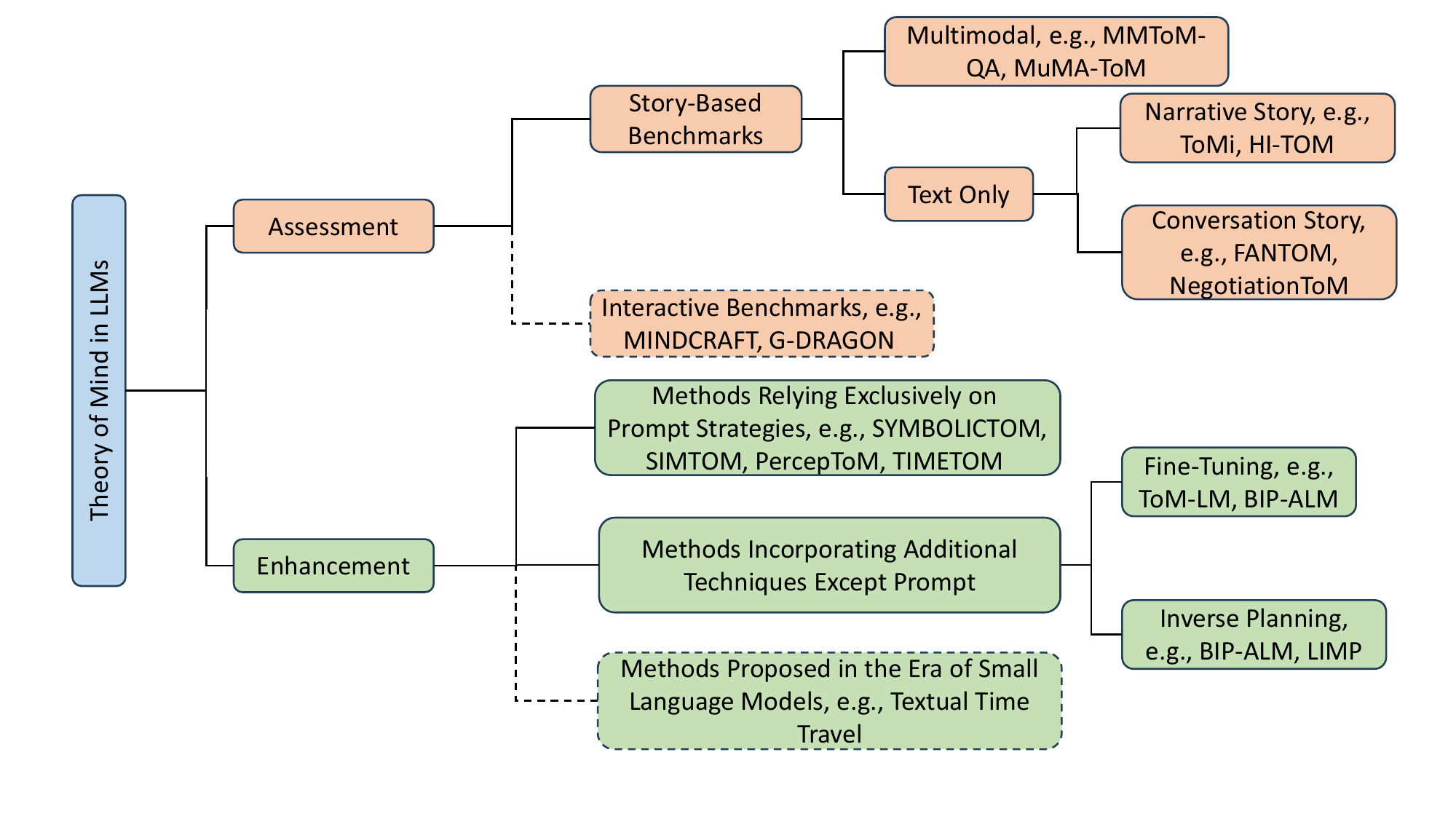}
    \caption{This paper reviews both the evaluation and enhancement of theory of mind capabilities in LLMs. For evaluation, we cover passive, story-based benchmarks and adaptable interactive benchmarks. For enhancement, we discuss recent effective methods, including prompt-only approaches and strategies incorporating additional techniques, as well as earlier methods from the era of small language models. Due to space constraints, interactive benchmarks and earlier methods (indicated in the dashed area) are provided in the Appendix \ref{interactive_benchmarks} and Appendix \ref{methods_in_small_language_models}.}
    \label{PaperScope}
\end{figure*}

To evaluate the ToM capabilities of LLMs, various benchmarks have been developed. Although \citet{ma-etal-2023-towards-holistic} advocate for a situated evaluation approach \cite{DBLP:conf/emnlp/BaraCC21, pmlr-v162-sclar22a, 10.24963/ijcai.2023/330}, story-based benchmarks remain prevalent. Notably, these story-based benchmarks have primarily emerged over the past two years (2023–2024) and have undergone rapid iterations. 
However, to the best of our knowledge, only \citet{ma-etal-2023-towards-holistic} has reviewed ToM benchmarks, and their focus is limited to those published up to 2023. Moreover, recent evaluations indicate that LLMs still lack robust ToM capabilities \cite{wu-etal-2023-hi, ma-etal-2023-tomchallenges, kim-etal-2023-fantom, gandhi2023understanding, xu-etal-2024-opentom, DBLP:journals/corr/abs-2404-13627, chen-etal-2024-tombench, jin-etal-2024-mmtom, Shi_Ye_Fang_Jin_Isik_Kuo_Shu_2025}, which has spurred significant research into strategies for enhancing these abilities. Yet, a detailed summary of these newly proposed strategies is still lacking.

Consequently, as illustrated in Figure \ref{PaperScope}, our survey covers both the evaluation and enhancement of LLMs' ToM capabilities. Specifically, we highlight recently proposed and widely used story-based benchmarks from 2023–2024 that have not been covered in existing reviews, analyzing their evolution over time. In addition, we provide a detailed review of the latest methods designed to improve ToM performance in LLMs, aiming to lay a foundation for future research in this area. To the best of our knowledge, this is the first study to offer such an in-depth analysis of both the assessment and advancement of ToM capabilities in LLMs. Our main contributions are summarized as follows:

\begin{itemize}
    \item \textbf{Broad Survey}. This paper provides the first broad survey that addresses both the evaluation and enhancement of LLMs' ToM capabilities. It offers a detailed analysis of recently proposed story-based benchmarks and the strategies developed to improve LLM performance in this domain.
    \item \textbf{In-depth Analysis}. We provide a detailed analysis of related research, focusing on the evolution of story-based benchmarks and the development of strategies to enhance ToM capabilities. This analysis offers a clearer and more structured overview of the progress made in this field.
    \item \textbf{Future Directions}. Building on our analysis, we identify several promising directions for future research, addressing both benchmark development and strategy refinement.
\end{itemize}

\section{Theory of Mind}

The Abilities in Theory of Mind Space (ATOMS) \cite{beaudoin2020systematic} defines seven mental states within ToM: beliefs, intentions, desires, emotions, knowledge, percepts, and non-literal communications. Detailed descriptions of each are provided in Appendix \ref{atoms}. As illustrated in Table \ref{benchmarks_since_2023}, beliefs are the most extensively studied mental state in ToM. In the following, we briefly explain several key concepts—such as orders, true-belief, and false-belief—that will recur throughout this paper.

The term "orders" refers to the number of mental state attributions required to answer a given question \cite{wu-etal-2023-hi}. For example, the question "Where will Sally look for her marble?" is considered first-order, while "Where does Anne think Sally will look for her marble?" is classified as second-order. When Sally's belief does not align with the reality, the question is categorized as a false-belief question; conversely, if her belief is accurate, it is referred to as a true-belief question.

\paragraph{Evaluating ToM Capabilities in LLMs}
As shown in Table \ref{benchmarks_since_2023}, more than ten benchmarks have been recently proposed to assess the ToM capabilities of LLMs. The majority of these benchmarks are based on well-established psychological tests, including the Sally-Anne Test \cite{BARONCOHEN198537} and the Smarties Task \cite{Perner1987ThreeyearoldsDW, gopnik1988children}, both of which are discussed in detail in Appendix \ref{Psychology_tests}.


\paragraph{Enhancing LLMs' ToM Capabilities} Table \ref{ToMMethods} lists recent approaches to enhance LLMs' ToM capabilities, which we classify as either relying solely on prompt engineering or incorporating additional techniques. Most strategies currently focus on belief-related questions, and Section \ref{strategies} provides a detailed discussion.

\begin{table*}[htbp]
  \centering
  \resizebox{\textwidth}{!}{
    \begin{tabular}{lccccccc}
\hline
\multirow{2}{*}{Benchmarks}                                              & \multicolumn{7}{c}{Mental States}                                                                                                                                                \\
                                                                         & Beliefs                  & Intentions                & Desires                   & Emotions                  & Knowledge                 & Percepts & Non-literal Communication \\ \hline
ToMi \cite{le-etal-2019-revisiting}                     & \checkmark &                           &                           &                           &                           &          &                           \\ \hline

MindGames \cite{sileo-lernould-2023-mindgames}          & \checkmark &                           &                           &                           &                           &     \checkmark     &                           \\ \hline
HI-TOM \cite{wu-etal-2023-hi}                           & \checkmark &                           &                           &                           &                           &          &                           \\ \hline
TOMCHALLENGES \cite{ma-etal-2023-tomchallenges}         & \checkmark &                           &                           &                           &                           &          &                           \\ \hline
FANTOM \cite{kim-etal-2023-fantom}                      & \checkmark &                           &                           &                           &                           &          &                           \\ \hline
BigToM \cite{gandhi2023understanding}                   & \checkmark &                           &          \checkmark                 &                           &                           &     \checkmark     &                           \\ \hline
OpenToM \cite{xu-etal-2024-opentom}                     & \checkmark & \checkmark &                           &      \checkmark                     &                           &          &                           \\ \hline
NegotiationToM \cite{DBLP:journals/corr/abs-2404-13627} & \checkmark & \checkmark & \checkmark &                           &                           &          &                           \\ \hline
TOMBENCH \cite{chen-etal-2024-tombench}                        & \checkmark & \checkmark & \checkmark & \checkmark & \checkmark &          & \checkmark \\ \hline
COMMON-TOM \cite{soubki-etal-2024-views}                & \checkmark &                           &                           &                           &                           &          &                           \\  \hline
SimpleToM \cite{gu2024simpletom}                & \checkmark &                           &      \checkmark                     &                           &                        \checkmark   &     \checkmark     &                           \\ \hline \hline
MMToM-QA \cite{jin-etal-2024-mmtom}                            & \checkmark & \checkmark                    &                           &                           &                           &          &                           \\ \hline
MuMA-ToM \cite{Shi_Ye_Fang_Jin_Isik_Kuo_Shu_2025}                             & \checkmark & \checkmark              &                           &                           &                           &          &                           \\  
\hline
\end{tabular}%
    }
\caption{Benchmarks and their coverage of mental states. The mental state categories are derived from the ATOMS \cite{beaudoin2020systematic}. In this paper, we treat the "goals" referenced in MMToM-QA \cite{jin-etal-2024-mmtom} and MuMA-ToM \cite{Shi_Ye_Fang_Jin_Isik_Kuo_Shu_2025} as equivalent to "intentions" in ATOMS. Benchmarks are divided into two categories: text-only benchmark and multimodal benchmark.}

  \label{benchmarks_since_2023}%
\end{table*}%
\section{Benchmarks for Evaluating ToM Capabilities in LLMs}
As mentioned previously, story-based benchmarks remain a leading approach for evaluating LLMs' ToM performance. In this work, we focus on these benchmarks, categorizing them by whether they include multimodal inputs. To compare different benchmarks and highlight areas for improvement, we present the coverage of mental states in Table \ref{benchmarks_since_2023}. Following \citet{ma-etal-2023-towards-holistic} and \citet{chen-etal-2024-tombench}, we adopt the mental state categories defined by the ATOMS framework \cite{beaudoin2020systematic}.

\subsection{Text-Only Benchmarks}
In this subsection, we first provide a brief overview of commonly used and recently proposed benchmarks for evaluating the ToM capabilities of LLMs\footnote{Due to space constraints, some benchmark descriptions are provided in the Appendix \ref{Continual-text-benchmarks}.}. We then highlight key trends observed in the evolution of these benchmarks.
\subsubsection{Overview of Text-Only Benchmarks}
\textbf{ToMi} \cite{le-etal-2019-revisiting}. Building on ToM-bAbi \cite{nematzadeh-etal-2018-evaluating}, ToMi enhanced the data generation process to create a more balanced dataset across various story types, introducing more complexity with random distractors. It also proposed generating all question types for each story, covering reality, memory, first-order and second-order beliefs (for each character). \citet{le-etal-2019-revisiting} introduced the concept of "story accuracy", where a story is considered correctly answered only if all associated questions are answered correctly. \citet{zhou2023far} transform ToMi from inferring mental states to probing an agent's action decisions, introducing a new evaluation paradigm called Thinking for Doing (T4D). Expanding on ToMi, \citet{jung-etal-2024-perceptions} developed Percept-ToMi by incorporating characters' perceptions.
While ToMi is widely used in current evaluations \cite{sclar-etal-2023-minding, wilf-etal-2024-think, jung-etal-2024-perceptions, hou-etal-2024-timetom}, it still relies on synthetic data, which employs simplified natural language compared to real-world scenarios.

\textbf{HI-TOM} \cite{wu-etal-2023-hi}. Recognizing the importance of higher-order ToM in social interactions and addressing the limitations of existing datasets, which are largely restricted to second-order belief evaluation, HI-TOM includes five questions per story. These questions correspond to reasoning levels from zeroth (equivalent to reality questions) to fourth order.
Building on the approaches of \citet{nematzadeh-etal-2018-evaluating} and \citet{le-etal-2019-revisiting}, HI-TOM's stories and questions are generated automatically using templates inspired by the Sally-Anne Test \cite{BARONCOHEN198537}, with distractor sentences incorporated into the narratives. While agent communication is integrated into the story generation process, the format of the stories remains narrative in nature.

\textbf{TOMCHALLENGES} \cite{ma-etal-2023-tomchallenges}. \citet{ma-etal-2023-tomchallenges} identified that inconsistencies in evaluating LLMs' ToM capabilities might arise from variations in evaluation methods and prompts. To address this, they designed six distinct questions for each "order" of ToM reasoning about each narrative story, using templates that span three generation settings: fully-constrained, semi-constrained, and open-ended. The stories are adapted from classic tasks like the Sally-Anne test and Smarties tasks \cite{BARONCOHEN198537, Perner1987ThreeyearoldsDW, gopnik1988children}, covering up to second-order belief questions. To enhance grading efficiency, they proposed an auto-grader based on the GPT-4\footnote{ \url{https://platform.openai.com/docs/models/gpt-4}} model. However, the effectiveness of this automatic evaluation tool may depend on the model's own ToM capabilities.

\textbf{FANTOM} \cite{kim-etal-2023-fantom}. To minimize reporting bias \cite{10.1145/2509558.2509563}, and better align with real-world scenarios, FANTOM uses conversations generated by LLMs instead of narrative stories. These conversations revolve around topics such as pets, risk-taking, and personal growth. To detect "illusory ToM"\footnote{\citet{kim-etal-2023-fantom} defines "illusory ToM" as cases where a model correctly answers some questions but fails to answer others that require the same type of ToM reasoning.}, six types of questions are posed for each conversation, including belief-related questions (up to second-order) and conclusive questions such as "List all the characters who know...". In addition, belief-related questions are tested under two settings: one with full factual knowledge and the other with facts limited to the perspective of a specific character. The current conversations in FANTOM are limited to small talk on specific topics generated by LLMs and it lacks prior knowledge about the individuals involved. This restricts its ability to simulate real-world social scenarios. Building on FANTOM, \citet{jung-etal-2024-perceptions} annotated characters' perceptions to develop Percept-FANTOM.

\textbf{BigToM} \cite{gandhi2023understanding}. \citet{gandhi2023understanding} proposed using LLM-generated evaluations to assess ToM capabilities in LLMs. By populating causal templates, they created BigToM, a benchmark designed to test LLMs' ToM capabilities across three dimensions: inferring beliefs from percepts, inferring actions from percepts (without access to beliefs), and inferring beliefs from actions (without access to percepts).
The narratives in BigToM are limited to first-order belief evaluations; however, the benchmark incorporates both percepts and desires, and holds the potential for expansion to higher-order belief evaluations.

\textbf{OpenToM} \cite{xu-etal-2024-opentom}. The main novelty of OpenToM lies in two key aspects: first, its narratives assign characters distinct personality traits and intentions that drive their actions; second, its questions extend beyond first-order and second-order belief queries to include emotion-related questions. Additionally, OpenToM introduces "accessibility" questions to evaluate LLMs' understanding of social commonsense. OpenToM facilitates the evaluation of a broader range of mental states in ToM capabilities. However, its current design is limited to two roles—a mover and an observer—and does not fully capture the complexity of human emotions \cite{zhan-etal-2023-evaluating}.  

\textbf{NegotiationToM} \cite{DBLP:journals/corr/abs-2404-13627}. Based on the Belief-Desire-Intention (BDI) framework \cite{1987Intention}, NegotiationToM evaluates machine ToM in multi-round real-world negotiation dialogues, assessing mental states like desires, second-order beliefs, and intentions. Built on the CaSiNo dataset \cite{chawla-etal-2021-casino}, it is the first ToM benchmark in negotiation contexts but remains passive and may suffer from data contamination due to reuse of existing data.   

\textbf{TOMBENCH} \cite{chen-etal-2024-tombench}. As shown in Table \ref{benchmarks_since_2023}, TOMBENCH integrates additional tasks from psychological research, such as the Unexpected Outcome Test \cite{dyck2001autism}, and covers all mental states defined in the ATOMS framework \cite{beaudoin2020systematic}, except for percepts. To prevent data contamination, TOMBENCH was constructed entirely from scratch, without leveraging any pre-existing datasets. Furthermore, TOMBENCH is a bilingual dataset containing both Chinese and English information. All test samples in TOMBENCH follow a multiple-choice question answering format, which minimizes subjective judgment. However, the multiple-choice question answering format poses a challenge in determining whether LLMs truly comprehend the questions and answer them correctly, as evaluation is based solely on the final answer option.

\subsubsection{Trends in Text-Only Benchmarks}
An analysis of the aforementioned benchmarks reveals that these benchmarks have grown increasingly complex, aiming to better reflect real-world scenarios. Apart from expanding from English benchmarks to multilingual benchmarks \cite{chen-etal-2024-tombench}, the primary advancements can be summarized as follows:
\begin{itemize}
\item \textbf{Order}. Initially, benchmarks focused on first-order belief questions \cite{grant2017can}. Over time, they expanded to second-order belief questions, which most benchmarks now support \cite{le-etal-2019-revisiting, ma-etal-2023-tomchallenges}. Some even enable the evaluation of fourth-order belief reasoning \cite{wu-etal-2023-hi}.
\item \textbf{Dataset Generation}. With the widespread adoption of LLMs, dataset generation methods have evolved from exclusively template-based approaches \cite{le-etal-2019-revisiting} to LLM-generated datasets created through template filling \cite{gandhi2023understanding, xu-etal-2024-opentom}. Recently, concerns about data contamination have led some benchmarks to rely on manually constructed datasets built from scratch by humans \cite{chen-etal-2024-tombench}.
\item \textbf{Context}. To better align with real-world scenarios, the format of contextual information has shifted from narrative stories \cite{le-etal-2019-revisiting, ma-etal-2023-tomchallenges} to multi-turn conversations \cite{kim-etal-2023-fantom, DBLP:journals/corr/abs-2404-13627, soubki-etal-2024-views}. Beyond format changes, the content has evolved from commonly used psychological tests \cite{le-etal-2019-revisiting, ma-etal-2023-tomchallenges} to topics that are more closely related to everyday life \cite{kim-etal-2023-fantom}.
\item \textbf{Questions}. To enable robust evaluation, benchmark questions have become more diverse and complex, evolving from single question answering task\cite{le-etal-2019-revisiting} to multiple question answering tasks\cite{ma-etal-2023-tomchallenges, kim-etal-2023-fantom}. Additionally, the content of the questions has changed, for instance, asking about the beliefs of a group of people rather than a single individual \cite{sileo-lernould-2023-mindgames, kim-etal-2023-fantom}.
\item \textbf{Mental States}. Benchmarks have expanded from evaluating only beliefs \cite{le-etal-2019-revisiting} to covering a wider range of mental states \cite{DBLP:journals/corr/abs-2404-13627, chen-etal-2024-tombench}, including percepts, desires, and even social commonsense knowledge \cite{xu-etal-2024-opentom}.
\end{itemize}
\begin{table*}[htbp]
  \centering
    \begin{tabular}{lcrr}
    \toprule
    Approaches & \multicolumn{1}{l}{Order Tested} & \multicolumn{1}{l}{Fine-Tuning} \\
    \midrule
    SYMBOLICTOM \cite{sclar-etal-2023-minding}  & \multicolumn{1}{l}{Up to third-order} & \multicolumn{1}{l}{No} \\
    SIMTOM \cite{wilf-etal-2024-think}  & \multicolumn{1}{l}{Up to second-order}   & \multicolumn{1}{l}{No} \\
    PercepToM \cite{jung-etal-2024-perceptions} & \multicolumn{1}{l}{Up to second-order} & \multicolumn{1}{l}{No} \\
    TIMETOM \cite{hou-etal-2024-timetom}  & \multicolumn{1}{l}{Up to third-order}    &   \multicolumn{1}{l}{No}  \\
    \midrule
    ToM-LM \cite{10.1007/978-3-031-71170-1_20}  & \multicolumn{1}{l}{Up to second-order}  & \multicolumn{1}{l}{Yes} \\
    BIP-ALM \cite{jin-etal-2024-mmtom}&   \multicolumn{1}{l}{Only first-order}&  \multicolumn{1}{l}{Yes} \\
    LIMP \cite{Shi_Ye_Fang_Jin_Isik_Kuo_Shu_2025}  & \multicolumn{1}{l}{Up to second-order} & \multicolumn{1}{l}{No} \\
    \bottomrule
    
    \end{tabular}%
    \caption{Comparison of approaches for enhancing LLMs' ToM capabilities. This table categorizes various methods into two groups based on whether they rely solely on prompt strategies. It also compares these methods in terms of the complexity of supported belief questions and the need for fine-tuning.}
  \label{ToMMethods}%
\end{table*}

\subsection{Multimodal Benchmarks}
In addition to the aforementioned developments, some researchers have recognized that ToM reasoning goes beyond simple text comprehension, which has led to the introduction of several multimodal ToM benchmarks. 

\textbf{MMToM-QA} \cite{jin-etal-2024-mmtom}. MMToM-QA is the first multimodal ToM benchmark that integrates both video and text to depict a person's activities within a household environment, specifically in VirtualHome \cite{DBLP:conf/cvpr/PuigRBLWF018}. The benchmark defines seven question types, categorized into two broad categories: belief inference and goal inference. Each question offers two possible answers, with one being significantly more likely to be correct based on the information provided in the text and video. Currently, MMToM-QA focuses on a single character finding objects in a household scenario and does not address higher-order beliefs. 

\textbf{MuMA-ToM} \cite{Shi_Ye_Fang_Jin_Isik_Kuo_Shu_2025}. Building on videos generated in VirtualHome \cite{DBLP:conf/cvpr/PuigRBLWF018}, this benchmark introduces the first multimodal, multi-agent ToM evaluation focusing on mental reasoning in embodied interactions. It includes three types of questions: belief inference, social goal inference, and belief-of-goal inference (a form of second-order belief question). Each question presents three options, with one being the most likely to be correct. Unlike MMToM-QA \cite{jin-etal-2024-mmtom}, MuMA-ToM provides distinctly separate information available through specific modalities, enabling a deeper understanding of how models integrate multimodal inputs to accurately answer each question. The text input is presented in two forms: conversations between two agents in the video or textual descriptions of half of the video. MuMA-ToM is currently limited to three social goals and two agents. Additionally, as it relies on video content from VirtualHome, further efforts are required to enhance its sim-to-real transferability for real-world applications.

The main limitation of current multi-modal benchmarks is their focus on household settings and reliance on synthetic videos. Sim-to-real transfer strategies may be necessary to adapt approaches tested on synthetic videos for real-world applications \cite{Shi_Ye_Fang_Jin_Isik_Kuo_Shu_2025}. Alternatively, a multi-modal benchmark incorporating real-world videos and more diverse scenarios \cite{jin-etal-2024-mmtom} is needed to enhance applicability and robustness. 

All these datasets are considered "passive benchmarks" \cite{DBLP:journals/corr/abs-2404-13627}, with LLMs acting as passive observers rather than active agents \cite{ma-etal-2023-towards-holistic, li-etal-2023-theory}. Evaluating LLMs as active agents to test their dynamic ToM capabilities represents a promising avenue for future research \cite{ma-etal-2023-towards-holistic, chen-etal-2024-tombench}. In Appendix \ref{interactive_benchmarks}, we present several interactive benchmark candidates that may be valuable in this regard.
\begin{figure*}[t]
    \centering
    \includegraphics[width=2\columnwidth]{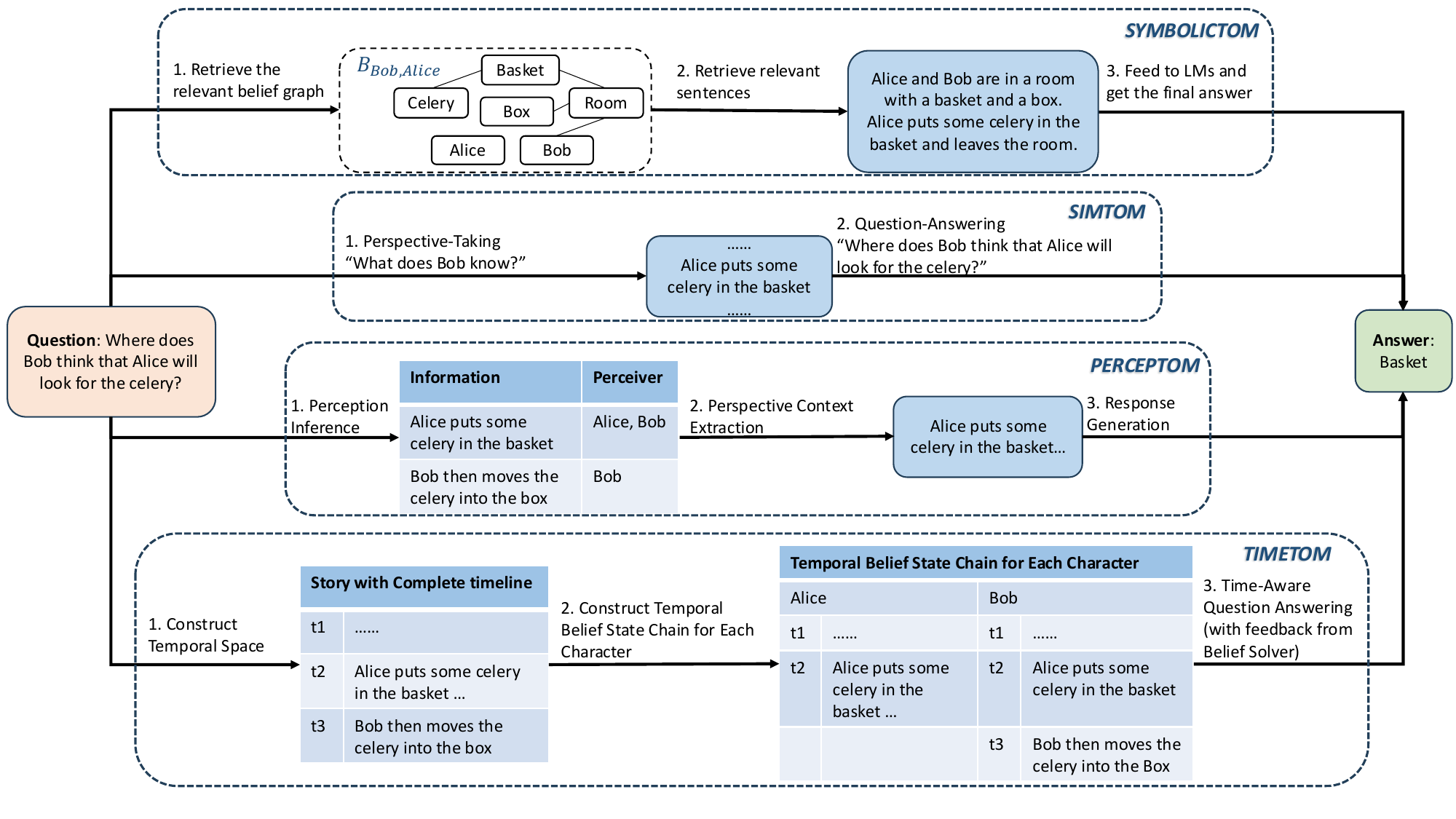}
    \caption{A comparison of methods for enhancing LLMs' ToM capabilities through different prompting techniques, using the following narrative as context. \textit{Story: Alice and Bob are in a room with a basket and a box. Alice puts some celery in the basket and leaves the room. Bob then moves the celery into the box \cite{sclar-etal-2023-minding}.}}
    \label{Comparison_between_prompting_methods}
\end{figure*}
\section{Strategies for Enhancing the ToM capabilities of LLMs} \label{strategies}
Most evaluations on these benchmarks highlight the limitations of LLMs in ToM capabilities \cite{wu-etal-2023-hi, ma-etal-2023-tomchallenges, kim-etal-2023-fantom, gandhi2023understanding, xu-etal-2024-opentom, DBLP:journals/corr/abs-2404-13627, chen-etal-2024-tombench, jin-etal-2024-mmtom, Shi_Ye_Fang_Jin_Isik_Kuo_Shu_2025}, prompting a surge in strategies aimed at improving their performance. Table \ref{ToMMethods} summarizes several recently proposed methods for enhancing the ToM capabilities of LLMs. We categorize these approaches into two classes: methods that rely solely on prompt strategies and methods that incorporate additional techniques, such as fine-tuning.


\subsection{Methods Relying Exclusively on Prompt Strategies}
Figure \ref{Comparison_between_prompting_methods} compares various approaches by illustrating how each method processes a second-order belief question.

\textbf{SYMBOLICTOM} \cite{sclar-etal-2023-minding}. As shown in Figure \ref{Comparison_between_prompting_methods}, they construct a belief graph for each character, capturing both their own beliefs and their beliefs about others, such as $B_{Bob}$ and $B_{Bob, Alice}$. During inference, entities mentioned in the question are identified, and the corresponding belief graph is located. Relevant sentences from this belief graph are retrieved and fed into the LLM for answering the question. Theoretically, it is capable of handling belief-related questions of any order. Experimental results on the variants of ToMi \cite{le-etal-2019-revisiting} demonstrated the effectiveness of this approach in addressing belief questions up to the third order. However, this method has several limitations. First, memory requirements grow exponentially with the order of belief-related questions. Second, the quality of the belief graph is critical; efficient and accurate methods for constructing belief graphs are essential as story complexity increases \cite{hou-etal-2024-timetom, jung-etal-2024-perceptions}. Additionally, the belief graph may lose historical information, making it incapable of answering questions such as, "List the locations of object A between the start and end of this story for a specific character".

\textbf{SIMTOM} \cite{wilf-etal-2024-think}. Inspired by “Simulation Theory” \cite{goldman2006simulating}, SIMTOM introduces a two-stage prompting framework that incorporates “Perspective Taking” \cite{goldman2006simulating} as an intermediate step. Briefly, SIMTOM involves identifying the events within a story that a specific character is aware of and using this filtered scenario to prompt LLMs for answers. SIMTOM enhances LLMs' capabilities in ToM tasks, and if the perspective-taking step were executed perfectly, current models could nearly solve existing benchmarks such as ToMi \cite{le-etal-2019-revisiting} and BigToM \cite{gandhi2023understanding}. However, the perspective-taking step often falls short of achieving human-level performance. Additionally, identifying which character's perspective the model should adopt becomes increasingly challenging as the complexity of the questions increases. These limitations may hinder SIMTOM's ability to handle higher-order ToM reasoning effectively \cite{hou-etal-2024-timetom}.

\textbf{PercepToM} \cite{jung-etal-2024-perceptions}. 
PercepToM is a framework designed to enhance the ToM capabilities of LLMs through a three-stage process. First, it identifies the perceiver of each unit of information. Second, it extracts and concatenates the units of information for which the perceiver includes the target character mentioned in the question. Finally, it prompts the LLMs with this curated information to answer the question. Experiments on ToMi \cite{le-etal-2019-revisiting} and FANTOM \cite{kim-etal-2023-fantom} have demonstrated its effectiveness. However, determining the target character in complex questions remains a challenge and requires further investigation, as it is crucial for providing the most relevant perspective context.  

\textbf{TIMETOM} \cite{hou-etal-2024-timetom}. TIMETOM enhances story comprehension by incorporating a timeline into the sentences of a story. It then identifies the sentences known to each character, forming what they call a “temporal belief state chain (TBSC)”. These sentences are further divided into "self-world beliefs" and "social-world beliefs".
For first-order questions, TIMETOM prompts LLMs using self-world beliefs to derive answers. For higher-order questions, it initially prompts LLMs with all sentences the character is aware of and refines the response using feedback from the Time-Aware Belief Solver. This tool transforms higher-order reasoning into first-order reasoning by intersecting the TBSCs of different characters.
Experiments on ToMi \cite{le-etal-2019-revisiting}, BigToM \cite{gandhi2023understanding} and FANTOM \cite{kim-etal-2023-fantom} demonstrate that TIMETOM effectively handles higher-order belief reasoning and achieves superior performance. Its innovative approach of transforming higher-order reasoning into first-order reasoning through temporal belief intersections is particularly noteworthy. However, a significant limitation is that most models currently struggle to accurately construct the temporal belief state chain for each character. 

All of these strategies take different approaches to identify the perceptions of the target character first, and then prompt LLMs to generate answers based on a limited story. They have demonstrated effectiveness across various benchmarks. Compared to fine-tuning-based methods, these strategies avoid costly training steps and have shown greater robustness to out-of-distribution samples \cite{sclar-etal-2023-minding}. However, all of these methods adopt a pipeline approach \cite{wan-etal-2023-joint} to solve ToM reasoning, which may lead to error propagation. This makes the initial step particularly critical. For instance, TIMETOM depends on generating the correct TBSC \cite{hou-etal-2024-timetom}. Despite their strengths, most LLMs still require significant improvement to effectively accomplish these tasks.

\subsection{Methods Incorporating Additional Techniques}
The main contribution of the following methods is their use of techniques beyond prompting, such as fine-tuning \cite{10.1007/978-3-031-71170-1_20, jin-etal-2024-mmtom} and inverse multi-agent planning \cite{Shi_Ye_Fang_Jin_Isik_Kuo_Shu_2025}.

\textbf{ToM-LM} \cite{10.1007/978-3-031-71170-1_20}. Inspired by \citet{olausson-etal-2023-linc, pan-etal-2023-logic, schick2024toolformer}, \citet{10.1007/978-3-031-71170-1_20} introduced ToM-LM, a framework that leverages LLMs for semantic parsing \cite{jia-liang-2016-data}, using one-shot learning to convert a ToM problem described in natural language into a symbolic formulation, and then evaluates this formulation using an SMCDEL model checker \cite{DBLP:journals/logcom/BenthemEGS18}. To enhance semantic parsing accuracy, the authors fine-tuned the tested LLM using <natural language, symbolic formulation> pairs. Integrating a model checker into this process adds transparency and verifiability to the approach. However, this method has two key limitations: preparing the fine-tuning pairs requires substantial logic expertise and effort, and the approach may not generalize well to open-ended questions.  


\textbf{BIP-ALM} \cite{jin-etal-2024-mmtom}. Bayesian Inverse Planning Accelerated by Language Models (BIP-ALM) extracts information such as the initial state and actions from videos and textual context, as well as hypotheses (including goals and beliefs) from questions and answer choices. All extracted information is represented in a symbolic format. Drawing inspiration from research on LLMs for decision-making \cite{pmlr-v162-huang22a, 10.5555/3600270.3602532}, the authors prompt LLMs with symbolic information—such as state, goal, and belief—to estimate the likelihood of observed actions. To further enhance the performance of LLMs in this task, they fine-tune the models using inputs comprising state, belief, and goal at specific timestamps to predict the corresponding actions. Using GPT-J \cite{wang2021gpt} and LLaMA 2 \cite{touvron2023llama} as base models, BIP-ALM achieved superior performance on the MMTOM-QA dataset. However, the MMTOM-QA dataset is currently limited to scenarios involving a character searching for objects in household environments, and BIP-ALM cannot support handling open-questions with the current setting.

\textbf{LIMP} \cite{Shi_Ye_Fang_Jin_Isik_Kuo_Shu_2025}. Inspired by BIP-ALM \cite{jin-etal-2024-mmtom}, Language Model-based Inverse Multi-agent Planning (LIMP) utilizes Vision-Language Models (VLMs) to extract information from videos and LLMs to extract information from textual context. These extracted details are then fused by an LLM. Mental state hypotheses—including beliefs of states, social goals, and beliefs about other agents’ goals—are derived from questions and answer choices and subsequently used as inputs for inverse multi-agent planning \cite{10.5555/2984093.2984303, netanyahu2021phase}. For the inverse multi-agent planning process, LIMP prompts an LLM with hypotheses derived from questions and multi-modal inputs to estimate action and utterance policies. Designed for multi-agent scenarios, LIMP outperforms BIP-ALM on the MuMA-ToM benchmark. Moreover, it demonstrates better generality by representing all information in natural language. However, hallucinations generated by VLMs in action recognition remain a significant source of errors in LIMP.

All the above strategies also approach ToM reasoning by breaking the problem into multiple steps. When symbolic representations are involved, fine-tuning is typically required to achieve better performance. However, a key limitation of TOM-LM, BIP-ALM, and LIMP is that they have only been evaluated on multiple-choice settings. Adapting these methods to handle open-ended question answering remains an open challenge that requires further exploration.  

\section{Future Directions}
Most evaluations indicate that LLMs still lack robust ToM reasoning abilities. To address this, future work should focus on developing better strategies and building more datasets—both for evaluation and fine-tuning. In this section, we first outline shared future directions for benchmarks and enhancement strategies, then discuss those specific to either goal\footnote{Due to space limitations, additional future directions are discussed in the Appendix \ref{future_work}.}.
Common future directions for both benchmarks and strategies involve:
\begin{itemize}
    \item \textbf{Expanding the scope of mental states}. As shown in Table \ref{benchmarks_since_2023} and Table \ref{ToMMethods}, current benchmarks and strategies mainly focus on belief-related reasoning, while other mental states require more attention \cite{ma-etal-2023-towards-holistic}. Future benchmarks should explore a broader range of mental states. To better evaluate LLMs' ToM capabilities, a wider variety of psychological tests should be integrated into the benchmarking process. \citet{FU2023101061} compiled a list of 127 tests designed to measure the ToM capabilities of children from birth to 12 years old, which could be further adapted for testing LLMs as the technology advances.
    \item \textbf{Addressing multi-modal ToM reasoning}. Humans interact with the world through multiple channels—an aspect that simple text-based stories cannot fully capture \cite{liu2023visual}. Research on ToM capabilities in LLMs aims to develop models that function as agents with robust ToM skills, potentially achieving human-level competence \cite{ma-etal-2023-towards-holistic}. Achieving this goal requires the ability to infer mental states from visual, auditory, contextual, and other cues. For instance, incorporating multimodal content such as short films \cite{dziobek2006introducing} or cartoons \cite{vollm2006neuronal, DBLP:conf/nips/ParmarP0LCTF24} into evaluation benchmarks could enhance ToM assessments. Moreover, current methods primarily focus on multiple-choice question answering in household contexts, and converting video information to text may lead to the loss of crucial details. Developing strategies that effectively handle multimodal inputs and preserve comprehensive information remains an important direction for future research.
    \item \textbf{Active benchmarks and strategies for agentic decision-making}. Passive benchmarks \cite{ma-etal-2023-towards-holistic, DBLP:journals/corr/abs-2404-13627} are insufficient. More research is needed to leverage LLMs as agents capable of making decisions in complex environments \cite{ma-etal-2023-towards-holistic, li-etal-2023-theory, zhou2023far, zhou-etal-2023-cast}, enabling deeper investigation of their ToM capabilities.
    
\end{itemize}

Next, we outline directions for further investigation to enhance LLMs' ToM capabilities:
\begin{itemize}
    \item \textbf{Exploring joint approaches}. Most existing methods rely on pipeline architectures with no feedback loops between stages. Error propagation is a significant issue in pipeline approaches \cite{yang-mitchell-2016-joint,liu-etal-2018-jointly}. For example, in SYMBOLICTOM, if the belief graph for a character is incorrect, it becomes difficult to obtain the correct answer when the question relates to this flawed belief graph. However, if conflicts are detected during the inference stage and feedback is provided during the belief graph construction stage, the accuracy of the inference and the coherence of the results can be improved. Investigating joint or iterative approaches that integrate feedback to improve ToM reasoning remains an open research question.
    \item \textbf{Progressive learning strategies for tackling ToM task complexity}. Given that ToM tasks can vary significantly in complexity, a continual/curriculum learning \cite{chen2018lifelong, soviany2022curriculum, 10444954} strategy may be needed to address these challenges progressively, starting with simpler tasks and advancing to more complex ones.
\end{itemize}

Finally, we believe that evaluating reasoning processes is essential. Assessing answer correctness alone is insufficient—datasets should be designed to support reasoning evaluation \cite{kawabata-sugawara-2023-evaluating, jung-etal-2024-perceptions, xu-etal-2024-reasoning}, calling for more effective and automated evaluation strategies. Likewise, probing a model’s internal representations of mental states \cite{bortoletto2024benchmarking} is an intriguing direction that warrants further exploration.

\section{Conclusion}
In this paper, we have conducted a detailed analysis of ToM research in relation to LLMs, exploring various benchmarks and strategies that aim to evaluate and enhance their ToM capabilities. Despite the introduction of numerous ToM benchmarks, achieving consistent results in assessing LLMs' ToM capabilities remains a challenge. This difficulty largely stems from the intrinsic complexity of ToM, which cannot be fully captured through a limited set of questions. Nevertheless, we remain optimistic that the development of more comprehensive benchmarks will support accurate evaluation of LLMs' true capabilities and further enhance their ToM performance. We hope that our paper will serve as an essential resource for those new to this field and that it will stimulate further research and advancements in understanding and improving the ToM capabilities in LLMs.
\section*{Limitations}
Theory of Mind (ToM) is an intriguing topic that has aroused interest across various domains, including psychology, reinforcement learning, and natural language processing. In this paper, since our focus is the theory of mind in large language models (LLMs), we structure our discussion following the evolution of LLMs—from the text-only paradigm to the multimodal paradigm. Regarding evaluation benchmarks, we primarily analyze story-based ToM benchmarks \cite{ma-etal-2023-towards-holistic}, including those that are either recently proposed or widely adopted to assess the effectiveness of various strategies. We do not delve into benchmarks that involve purely spatial scenarios \cite{pmlr-v162-sclar22a} (e.g., BIB \cite{gandhi2021baby}). Concerning strategies, we focus on analyzing the currently proposed and effective methods tested with LLMs. However, we also present several interactive benchmarks and strategies developed prior to the widespread adoption of LLMs in Appendix \ref{interactive_benchmarks} and Appendix \ref{methods_in_small_language_models}, which have the potential to be adapted for research in this area.

\section*{Acknowledgments}
We sincerely thank all reviewers for their valuable feedback. This work was supported by A*STAR CRF funding awarded to Cheston Tan.
\bibliography{acl_latex}

\appendix
\clearpage
\newpage
\section{Psychological Tests} \label{Psychology_tests}
\subsection{Sally-Anne Test}

\begin{figure}[t]
    \centering
    \includegraphics[width=\columnwidth]{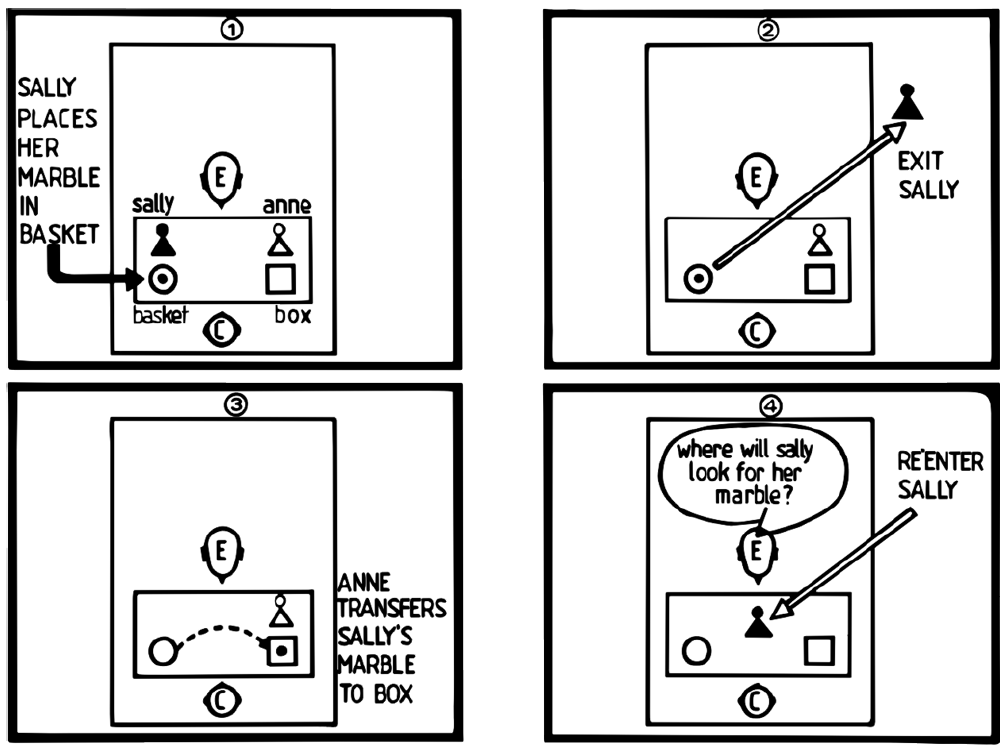}
    \caption{Experimental scenario in Sally-Anne test \cite{BARONCOHEN198537}.}
    \label{Sally-Anne Test}
\end{figure}

The Sally-Anne test \cite{BARONCOHEN198537} has been widely used in psychology studies. As illustrated in Figure \ref{Sally-Anne Test}, the test typically involves two characters, Sally and Anne. While Sally is away, Anne hides a marble, setting the stage for a set of questions designed to assess children's ToM capabilities.
\begin{itemize}
    \item Belief Question: "Where will Sally look for her marble?"
    \item Reality Question: "Where is the marble really?"
    \item Memory Question: "Where was the marble in the beginning?"
\end{itemize}

\subsection{Ice Cream Van Experiment}
The Ice Cream Van Experiment \cite{PERNER1985437} features three characters: Mary, John, and the ice cream man. In this scenario, the story unfolds as follows:
\begin{itemize}
    \item Mary and John both see the ice cream man in the park, and he informs them that he will be there for the entire afternoon.
    \item Mary leaves the park and goes home. Later the ice cream man departs and tells John that he is heading to church.
    \item On his way to the church, he runs into Mary and informs her that he will be selling ice cream near the church for the rest of the afternoon.
\end{itemize}

Based on this setup, researchers can pose both first-order and second-order belief questions. For example:
\begin{itemize}
    \item First-Order Belief Question: “Does Mary know the van is at the church?”
    \item Second-Order Belief Question: “Where does John think Mary will go to buy ice cream?”
\end{itemize}

\subsection{Smarties Task}

The Smarties task \cite{Perner1987ThreeyearoldsDW}, sometimes referred to as "unexpected contents"\footnote{\url{https://en.wikipedia.org/wiki/Theory\_of\_mind}} or "appearance-reality"\footnote{\url{http://www.sfu.ca/~hedberg/ToM.pdf}} task, involves asking children what they think is inside a box that looks like it contains smarties. After the child guesses "smarties", they are shown that the box actually holds a different item, such as pencils. The box is then resealed, and the child is asked what they think someone else, who hasn't seen the real contents, will believe is inside. The child succeeds if they say the other person will think "smarties" are in the box, and fails if they say the other person will think it contains pencils. According to Gopnik and Astington's research, children generally pass this task by the age of four or five \cite{gopnik1988children}. 

These task, with adjustments to elements like the characters, the container, or the objects involved, forms the basis for various ToM benchmarks \cite{ma-etal-2023-tomchallenges}.

\section{Abilities in Theory of Mind Space (ATOMS)} \label{atoms}
The ATOMS framework \cite{beaudoin2020systematic} defines seven key mental states essential for Theory of Mind: beliefs, intentions, desires, emotions, knowledge, percepts, and non-literal communications. Here’s a brief explanation of each:
\begin{itemize}
    \item Beliefs: Represent an individual’s assumptions or interpretations of reality.
    \item Intentions: Capture the planned actions or motivations behind behavior.
    \item Desires: Reflect what an individual wants or hopes to achieve.
    \item Emotions: Encompass the range of feelings a person experiences.
    \item Knowledge: Consists of the factual information and experiences a person has accumulated.
    \item Percepts: Involve the sensory information or observations made by an individual.
    \item Non-literal Communications: Cover the use of figurative language, sarcasm, and implied meanings that go beyond the literal interpretation of words.

\end{itemize}
Each of these mental states plays a distinct role in how individuals understand and predict others' behavior during social interactions.

\section{Overview of Additional Text-Only Benchmarks} \label{Continual-text-benchmarks}

\textbf{ToM-bAbi} \cite{nematzadeh-etal-2018-evaluating}. Drawing inspiration from the influential Sally-Anne task \cite{WIMMER1983103, BARONCOHEN198537} and adopting the dataset generation procedure of the bAbI \cite{DBLP:journals/corr/WestonBCM15} dataset generation procedure, \citet{grant2017can} took the initial step in designing benchmarks aimed at evaluating the mental-state reasoning abilities of question answering models, specifically focusing on first-order beliefs \cite{nematzadeh-etal-2018-evaluating}. Building upon this foundation and drawing inspiration from the "beliefs about belief" questions posed in the Ice Cream Van Experiment \cite{PERNER1985437}, \citet{nematzadeh-etal-2018-evaluating} introduced two new datasets, ToM and ToM-easy, that further enables the assessment of a model's capacity to reason about Second-order False Belief. These benchmarks were later referred to as ToM-bAbi by \citet{le-etal-2019-revisiting}.

\textbf{MindGames} \cite{sileo-lernould-2023-mindgames}. Unlike most of benchmarks, MindGames incorporates variations of the Muddy Children and Drinking Logicians problems \cite{DBLP:books/daglib/p/Eijck14}. During data generation, dynamic epistemic logic \cite{DBLP:books/daglib/p/Eijck14} problems are first created and then verbalized using predefined templates. Although the published MindGames dataset is currently limited to testing second-order beliefs, it has the potential to assess higher-order reasoning.

In MindGames, some questions involve evaluating hypotheses like: "Catherine can now know whether Shelley can know whether or not everyone's forehead is muddy"\footnote{\url{https://huggingface.co/datasets/sileod/mindgames}}. Such problems are more challenging because they require understanding and integrating information from all individuals involved. Additionally, with its use of visual cues \cite{chen-etal-2024-tombench}, MindGames also explores the "percept" dimension of mental states, further broadening its evaluation capabilities.

\textbf{COMMONTOM} \cite{soubki-etal-2024-views}. COMMON-TOM, grounded in the concept of common ground (CG) \cite{WILKESGIBBS1992183, stalnaker2002common}, is a question-answering benchmark designed around spoken dialogues. It evaluates the ToM capabilities of LLMs, particularly their ability to handle belief-related questions up to the third order.

\textbf{SimpleToM} \cite{gu2024simpletom}. SimpleToM consists of two-sentence narrative stories, each accompanied by three binary questions designed to assess both "explicit ToM" \cite{gu2024simpletom} and "applied ToM" \cite{gu2024simpletom} capabilities of LLMs. For "explicit ToM," the questions require models to predict a person’s mental state, while for "applied ToM," they involve predicting a person’s behavior or determining whether a behavior is reasonable. The stories are based on ten real-life scenarios, such as "selecting a food item in a grocery store" and "seeking provider information in healthcare." As shown in Table \ref{benchmarks_since_2023}, the benchmark supports the evaluation of various mental states, including beliefs, desires, perceptions, and knowledge. Although some scenarios involve two individuals, the belief-related questions focus exclusively on first-order reasoning.

\section{Comparative Information on Benchmarks}
In addition to the comparison in Table \ref{benchmarks_since_2023}, we present Table \ref{more_comparison} for a more detailed comparison, allowing readers to gain a deeper understanding of each benchmark, including aspects such as story structure, QA formats, and more.
\begin{table*}[htbp]
  \centering
  \renewcommand\arraystretch{1.5}
  \setlength{\tabcolsep}{4mm}{
  \resizebox{\textwidth}{!}{
    \begin{tabular}{p{7em}p{8.275em}p{4.145em}p{6.59em}rp{5.68em}p{6.455em}}
    \toprule
    Benchmarks & Databases & Story & QA Formats & \multicolumn{1}{p{3em}}{Highest Order} & Language & Publish \\
    \midrule
    ToMi  & Sally-Anne Test & Narrative & Open QA & 2     & English & EMNLP-IJCNLP (2019) \\
    MindGames & Muddy Children and Drinking Logician problems & Narrative & Yes/No QA & 2     & English & EMNLP Findings (2023) \\
    HI-TOM & Sally-Anne Test & Narrative & Multi-Choice QA & 4     & English & EMNLP Findings (2023) \\
    TOMCHALLENGES & Sally-Anne and Smarties Tests & Narrative & Fill-in-the-Blank, Multi-Choice, Yes/No, Open-Ended QA etc. & 2     & English & CoNLL (2023) \\
    FANTOM & LLM-generated conversations related to pets, personal growth, and travelling etc. & Conversation & Open-Ended, Yes/No, Multi-Choice QA & 2 & English & EMNLP (2023) \\
    BigToM & LLM-generated based on causal templates & Narrative & Multi-Choice QA (2 answer options) & 1     & English & NeurIPS Track on Datasets and Benchmarks (2023) \\
    OpenToM & LLM-generated based on Sally-Anne Test with personality traits and preferences & Narrative & Multi-Choice QA (2 or 3 answer options) & 2     & English & ACL (2024) \\
    NegotiationToM & CaSiNo & Conversation & Multi-Choice QA, Ranking & 2     & English & EMNLP Findings (2024) \\
    TOMBENCH & Build from scratch & Narrative & Multi-Choice QA & 2     & English, Chinese & ACL (2024) \\
    COMMON-TOM & Common ground (CG) Corpus & Conversation & Yes/No QA & 3     & English & ACL Findings (2024) \\
    SimpleToM & LLM-generated based on ten real-life scenarios & Narrative & Yes/No QA & 1     & English & arXiv (2024) \\
    MMToM-QA & People looking for objects scenario & Narrative & Multi-Choice QA (2 answer options) & 1     & English & ACL (2024) \\
    MuMA-ToM & Household activities & Narrative and Conversation & Multi-Choice QA (3 answer options) & 2     & English & AAAI (2025) \\
    \bottomrule
    \end{tabular}%
    }}
    \caption{Comparison of benchmarks based on their foundations, story and QA Format, highest order of reasoning involved, language coverage, and publication details. Due to width constraints, citations have been omitted from this table.}
  \label{more_comparison}%
\end{table*}%

\section{Interactive Benchmarks} \label{interactive_benchmarks}
Although they are not yet widely used for evaluating LLMs’ ToM capabilities, a few interactive benchmarks or tasks have been proposed to help improve or test neural models’ ToM skills. Most of these benchmarks are derived from game-based or reinforcement learning (RL) research scenarios, such as Minecraft \cite{DBLP:conf/emnlp/BaraCC21}, Dungeons and Dragons \cite{zhou-etal-2023-cast}, and Grid \cite{pmlr-v80-rabinowitz18a, pmlr-v162-sclar22a, ma-etal-2023-towards-holistic}. These benchmarks can be adapted to develop situated and interactive evaluations for LLMs. A few potential interactive benchmark candidates are listed below.

\textbf{MINDCRAFT} \cite{DBLP:conf/emnlp/BaraCC21}. It is set in a Minecraft\footnote{\url{https://www.minecraft.net/en-us}} environment where two partners collaborate to create new structures by combining blocks. Each partner possesses asymmetric knowledge (represented in a knowledge graph) and distinct skill sets, requiring them to negotiate via a text channel to effectively achieve their final goal.
Regarding visual information, both first-person and third-person perspectives are available. The dataset includes three types of questions: task completion status, player knowledge, and the player’s current task. Unlike story-based benchmarks, MINDCRAFT places a greater emphasis on perception and knowledge aspects within the ToM.

\textbf{SymmToM} \cite{pmlr-v162-sclar22a}. SymmToM is a symmetric multi‐agent environment in which agents can speak, listen, see one another, and navigate freely across a grid world. All agents—each possessing identical abilities and roles—actively participate in a information‐gathering game. Although agents have perfect vision of the entire grid, their hearing is confined to a limited range. To gather information efficiently, agents should model one another’s mental states. Ultimately, each agent’s goal is to maximize its reward by collecting all available information.

\textbf{Generating Guidance in Goal-Driven and Grounded Communication (G4C)} \cite{zhou-etal-2023-cast}. In the G4C task, a dataset called G-DRAGON is created, based on a dialogue dataset set in the Dungeons and Dragons scenario \cite{callison-burch-etal-2022-dungeons}. This task explicitly incorporates the intent of the Dungeon Master into the natural language generation process to examine whether integrating intent and ToM enhances the communicative abilities of computational models.

\textbf{Situated ToM in Grid World} \cite{ma-etal-2023-towards-holistic}. \citet{ma-etal-2023-towards-holistic} advocate for a situated evaluation of ToM, as it encompasses more aspects than text-only datasets while also mitigating data contamination and shortcuts. They compile a multiple-choice question-answering dataset in MiniGrid \cite{chevalier2018minimalistic}, which covers all aspects of ATOMS as well as a "reality check."

Beyond these benchmarks, there is still room for further improvement. For example, incorporating more agents, as the current MINDCRAFT \cite{DBLP:conf/emnlp/BaraCC21} and G-DRAGON \cite{zhou-etal-2023-cast} benchmarks involve only two; integrating multimodal information, as G-DRAGON \cite{zhou-etal-2023-cast}, and Situated ToM in Grid World \cite{ma-etal-2023-towards-holistic} currently focus solely on text input; and extending the method from handling introspective beliefs to addressing first-order or even higher-order beliefs questions \cite{li-etal-2023-theory}. Additionally, negotiation-based or tabletop games, such as auction games, and human-AI collaboration tasks \cite{zhang2024mutual}, could further contribute to benchmark development for evaluating the ToM capabilities of LLMs.

\section{Methods Proposed in the Era of Small Language Models} \label{methods_in_small_language_models}
Before the widespread adoption of LLMs, researchers had already recognized the importance of theory of mind and proposed several methods to enhance models’ ToM capabilities. For example, 

\textbf{Textual Time Travel} \cite{arodi-cheung-2021-textual-time}. This framework consists of two stages: a memory updater and a textual time-travel mechanism. The memory updater captures a snapshot of the memory contents at every time step. The textual time-travel mechanism then identifies a relevant time step, retrieves the corresponding memory contents, and uses them to generate an answer to a given query.
They introduce both a heuristic (rule-based) and a learning-based version of the textual time-travel mechanism, guided by three key assumptions:
\begin{itemize}
    \item Local entity perception: An entity is only aware of objects and other entities in its current location, with no knowledge of those elsewhere.
    \item Recency assumption: An entity assumes that the most recent information it has about an object or another entity is correct.
    \item Reciprocity assumption: Entities assume that all other entities also operate under the local perception and recency assumptions.
\end{itemize}

Additionally, multiple sequence-to-sequence models have been implemented \cite{DBLP:conf/emnlp/BaraCC21}, and several reinforcement learning approaches—such as Explicitly Modeling Symmetric Theory of Mind \cite{pmlr-v162-sclar22a} and ToM-Inspired RL \cite{zhou-etal-2023-cast}—have also been proposed. These methods could also guide future improvements in LLMs' ToM capabilities.

\section{Additional Future Directions} \label{future_work}
\textbf{Exploring multilingual benchmarks}. Currently, most of the benchmarks are still in English. Multilingual benchmarks \cite{chen-etal-2024-tombench} are crucial for evaluating and advancing LLMs' comprehensive ToM capabilities across different languages.

\textbf{Advancing higher-order ToM reasoning}. While most existing strategies and benchmarks remain focused on second-order beliefs, higher-order reasoning is crucial in contexts such as multi-party negotiations. As such, advancing higher-order ToM capabilities warrants greater attention and innovation \cite{wu-etal-2023-hi, hou-etal-2024-timetom} to achieve "advanced ToM" \cite{BIALECKAPIKUL2017145, sclar-etal-2023-minding} in LLMs.

\textbf{Exploring the ToM capabilities in smaller language models}. Most strategies have been tested on models with over 7 billion parameters. However, these strategies may not be effective on models with fewer than 7 billion parameters, as smaller language models typically have a reduced capacity for understanding instructions \cite{hou-etal-2024-timetom}. Additionally, LLMs require substantial resources, which may not always be available. Therefore, it is crucial to investigate the ToM capabilities of smaller language models. Appendix \ref{methods_in_small_language_models} presents several effective methods developed during the era of small language models that may offer valuable insights for this research direction.


\end{document}